\documentclass{ecai}

\usepackage{danudefs}
\usepackage{times}
\usepackage{url}
\usepackage{graphicx}
\usepackage{booktabs} 
\usepackage{caption} 
\usepackage{array,multirow}
\usepackage{xcolor}
\usepackage{colortbl} 
\usepackage{wrapfig}
\usepackage{caption}
\usepackage{xcolor}
\usepackage{tabularx}
\usepackage{url}
\usepackage{hyperref}

\usepackage{graphicx}

 \usepackage{todonotes} 
 
 \makeatletter
\if@todonotes@disabled

\else

\fi
\makeatother

 \newcommand\DB[2][]{\todo[inline, caption={2do}, #1]{
\begin{minipage}{\textwidth-4pt}\textbf{DB:~}#2\end{minipage}}}

\begin{document}

\title{Meta-Embedding as Auxiliary Task Regularization}

\author{James O' Neill \and Danushka Bollegala \institute{University of Liverpool,UK, Email: \{\texttt{james.o-neill, danushka\}@liverpool.ac.uk}}}

\maketitle
\bibliographystyle{ecai}

\begin{abstract}
Word embeddings have been shown to benefit from ensambling several word embedding sources, often carried out using straightforward mathematical operations over the set of word vectors. More recently, self-supervised learning has been used to find a lower-dimensional representation, similar in size to the individual word embeddings within the ensemble. However, these methods do not use the available manually labeled datasets that are often used solely for the purpose of evaluation. We propose to reconstruct an ensemble of word embeddings as an auxiliary task that regularises a main task while both tasks share the learned meta-embedding layer. We carry out intrinsic evaluation (6 word similarity datasets and 3 analogy datasets) and extrinsic evaluation (4 downstream tasks). For intrinsic task evaluation, supervision comes from various labeled word similarity datasets. 
Our experimental results show that the performance is improved for all word similarity datasets when compared to self-supervised learning methods with a mean increase of $11.33$ in Spearman correlation. 
Specifically, the proposed method shows the best performance in 4 out of 6 of word similarity datasets when using a cosine reconstruction loss and Brier's word similarity loss. 
Moreover, improvements are also made when performing word meta-embedding reconstruction in sequence tagging and sentence meta-embedding for sentence classification.
\end{abstract}

\section{Introduction}

Distributed word representations have shown to improve performance in numerous natural language processing (NLP) tasks~\cite{lai2015recurrent,kenter2015short,liu2016recurrent,turian2010word}. 
Given that the performance on intrinsic (e.g word similarity, analogy) and extrinsic (e.g sentiment analysis, dependency parsing, machine translation etc.) supervised tasks is dependent on the model used for producing word embeddings (e.g $\mathtt{skipgram}$, $\mathtt{cbow}$~\cite{mikolov2013efficient}), it is clear that each model exploits different aspects of the semantic space. 
The goal in meta-embedding learning~\cite{coates2018frustratingly,yin2015learning,bao2018meta,Bollegala:IJCAI:2018,kiela-etal-2018-dynamic} is to learn a single, (possibly lower-dimensional) common embedding space by combining multiple, pre-trained \emph{source} word embeddings, without re-training the sources or requiring access to the linguistic resources such as text corpora or dictionaries that were used to train the source embeddings.
Meta-embedding learning methods aim to be computationally and resource-wise efficient by not retraining source embeddings.

Existing approaches to meta-embedding rely on self-supervised learning such as autoencoding~\cite{bao2018meta} to find a lower-dimensional hidden representation of the set of source embedding (further discussed in \S~\ref{sec:related:meta-embedding}). 
This can be advantageous in cases where (1) pre-training is expensive, (2) pre-trained embeddings are available but not the algorithm or the training data used,
and (3) the available source embeddings vary in their dimensionalites. 
However, for problems that rely on representations that are better aligned with human judgment (e.g., \textit{genuine} similarity~\cite{hill2016simlex}), word embeddings and word meta-embeddings struggle to perform well when only given co-occurrence statistics. 
How to best incorporate task-specific human judgements into the meta-embedding learning process remains a challenging and unsolved problem.

To overcome this challenge, we propose a semi-supervised multi-task learning (MTL) approach that combines the benefits of self-supervised learning for finding a lower-dimensional representation of the concatenated word meta-embeddings, while also learning to perform inference on word similarity used as an intrinsic task, and extrinsic downstream tasks such as sequence tagging using a siamese network that incorporates a shared representation from an autoencoder (AE). 
We consider meta-embedding reconstruction as an auxiliary task in contrast to the main classification task, hence the reference to multi-task learning.
We evaluate our approach on held-out word similarity datasets and also include an evaluation on the transferability of the resultant word meta-embeddings on three analogy datasets.
We find that performance is improved for \emph{all} word similarity datasets with a mean increase of 11.33 points in the Spearman Correlation coefficient $\rho_s$, when compared to self-supervised learning methods. Below we summarize the main aspects of our work.   

\paragraph{Angular Cost Functions:}

In meta-embedding learning we use source embeddings trained on different resources, in which case it is important to keep the semantic orientation of words by preserving the angle between word embeddings, not only the length. Cosine similarity, a popularly used measure for computing the semantic relatedness among words, ignores the length related information. We also note the relationship between KL-divergence and cosine similarity in that both methods perform a normalization. 
Hence, we compare Mean Squared Error (MSE) and Mean Absolute Error (MAE), against KL-divergence and Squared Cosine similarity for the purpose of learning meta-embeddings and show that loss functions that account for vector orientation can outperform the former MSE and MAE objectives that only preserve length but not orientation.

\paragraph{Supervision from Manual Annotations:}
Currently, word embeddings and word meta-embedding methods do not exploit the available manual annotations in the learning process such as word similarity ratings. 
In particular, word vectors often struggle to preserve true similarity, which in many cases is difficult to identify from statistical associations alone. Hill et al.~\cite{hill2016simlex} found word embeddings to struggle for word similarity in comparison to word association, particularly for abstract concepts. Our method addresses this by learning to reconstruct meta-embeddings while sharing the hidden layer to jointly predict on a main task (e.g word similarity, NER or Sentiment classification). In the case of word similarity, we argue that this explicit use of true similarity scores can greatly improve embeddings for tasks that rely on true similarity. This is reflected in our results for Simlex~\cite{hill2015simlex} and rare word~\cite{luong2014addressing} datasets as we find 29.63 and 27.05 point increases in $\rho_s$ respectively. 

\paragraph{Dealing with Out-of-Vocabulary Words:}
Word vectors suffer in performance for out-of-vocabulary words that are not seen during training. 
This is an issue on evaluation datasets that gauge performance on words that are morphologically complex, rare~\cite{luong2013better} or are highly abstract conceptually ~\cite{hill2016simlex}. In fact, Luong et al.~\cite{luong2013better} have used sub-word vectors for such issues. Alternatively, Cao and Rei~\cite{cao2016joint} have used a character-level composition from morphemes to word embeddings where morphemes that yield better predictive power for predicting context words are given larger weights, showing improvements over word-based embeddings for syntactic analogy answering. Their model incorporated morphological information into character-level embeddings which in turn, produced better representations for unseen words. Word meta-embeddings allow for much larger coverage by combining the ensemble set of pre-trained word embeddings, trained from different corpora. 
This approach also allows for sub-word level combinations between embeddings.   

\paragraph{Motivation for Multi-Task Learning:}
Using shared representations in multi-task learning has led to better generalization performance on each respective task in prior work~\cite{caruana1998multitask}, 
(1) by introducing relevant inter-dependent features, 
(2) by regularizing a model using the performance on multiple tasks,
 (3) using future tasks to interpolate to present tasks, 
 (4) by improving the model's ability to learn general features from noisy signals,
  and (5) by potentially exploiting the loose structure among the parent tasks that aid more specific downstream child sub-tasks (e.g tasks are designated based on the word relation type such as hyponymy, antonymy or synonymy). 

\section{Related Work}
\label{sec:related}

\subsection{Word Meta-Embeddings}
\label{sec:related:meta-embedding}

The most straightforward approaches to meta-embeddings are: concatenation (CONC) and averaging (AV). The former is limited in the dimensionality size and the latter only preserves the mean of the embedding set. 
Coates et al.~\cite{coates2018frustratingly} showed that if the word vectors are approximately orthogonal, AV approximates CONC even though the embedding spaces may be different. Hence, we include AV in our comparisons of unsupservised methods.
Singular Value Decomposition (SVD) has been used to factorize the embeddings into a lower-rank approximation of the concatenated meta-embedding set. Yin et al.~\cite{yin2015learning} use of a projection layer for meta-embedding (known as 1\texttt{TO}N) optimized using an $\ell_2$-based loss. Similarly, Bollegala et al.~\cite{Bollegala:IJCAI:2018} have focused on finding a linear transformation between count-based and prediction-based embeddings, showing that linearly transformed count-based embeddings can be used for predictions in the localized neighborhoods in the target space. 
Kiela at al.~\cite{kiela-etal-2018-dynamic} proposed a dynamically weighted meta-embedding method for representing sentences where they first project each source embedding using a source-specific projection matrix to a common vector space where they can add the projected source embeddings multiplied by an attention weight. 
They consider contextualised word embeddings given by the concatenated forward and backward hidden states of a BiLSTM. 
The attention weights are learnt using labelled data for sentence-level tasks such as sentiment classification and textual entailment. 
On the other hand, in this paper, we consider meta-embedding of context-independent pre-trained source word embeddings.

Bao and Bollegala~\cite{bao2018meta} used three AE variants for meta-embeddings: (1) Decoupled Autoencoded Meta Embedding (DAEME) that keep activations separated for each respective embedding input during encoding and uses a reconstruction loss for both predicted embeddings while minimizing the mean of both predictions, (2) Coupled Autoencoded Meta Embedding (CAEME) that learn to predict from a shared encoding (dropping the expectation minimization loss used in DAEME), and (3) Averaged Autoencoded Meta-Embedding (AAME) simply use the average of the embedding set instead of concatenation. 
We consider those autoencoding methods in evaluations (see \S~\ref{sec:experiments}). 
We also propose two important variants of the aforementioned AEs in \S~\ref{sec:method}. 
The first variant predicts a target embedding from an embedding set using the remaining embedding set, whereafter training, the single hidden layer is used as the word meta-embedding. 
The second variant is similar to the first except an AE is used for each input embedding to predict the designated target embedding, followed by averaged pooling over the resulting hidden layers. 

\subsection{Multi-Task Learning Representations}

Ando and Zhang~\cite{ando2005framework} learn representations for multiple tasks in a partially supervised and unsupervised way, which draws similarities to the work presented in this paper. The challenge is characterized as (1) predicting labels for an auxiliary task from another task that is trained with full supervision, and (2) both tasks are in some way related. Both characteristics also hold for the work presented in our paper with the subtle difference that we are using many related word similarity datasets with full supervision to predict the auxiliary task (i.e a held-out word similarity dataset for testing). 
Part-of speech (PoS) tags of the contextual words are used to predict the current word's PoS tag in a self-supervised fashion, similar to masking used in word embedding learning~\cite{mikolov2013efficient,devlin2018bert}. 
Specifically, some features that are to be predicted for text categorization are masked out to create auxiliary prediction tasks.
Ando and Zhang~\cite{ando2005framework} used this method to obtain good performance on PoS tagging and Named Entity Recognition (NER) using a language model that predicts a target word given its context words.  
Similarly, Collobert et al.~\cite{collobert2011natural} proposed a unified neural network architecture for learning PoS tagging, NER, Chunking and Semantic Role Labeling all at once with parameter sharing using unlabeled training data.

Dong et al.~\cite{dong2015multi} used MTL to improve the quality of machine translation to multiple target languages. 
They share the source language representations in the encoder-decoder sequence model considering the availability of the required parallel data. 
Their model showed higher BLEU scores over independent sequence-to-sequence language models when there is full and partial availability of parallel data, highlighting the importance of integrating related source language representations.

Liu et al.~\cite{liu2015representation} used MTL for query classification using multiple binary classifiers, and web search ranking based on maximum likelihood with deep neural networks. Their MTL architecture consisted of three shared hidden layers that use character and word $n$-gram inputs, where the last layer is an independent task-specific layer for query classification and web search respectively. MTL showed large improvements over the baseline support vector machines and neural networks that learn each task independently.

All of the aforementioned work focus on using MTL on high-level natural language tasks. This work is distinct in that we use meta-embedding reconstruction as a regularization technique, treating it as an auxiliary task to the main intrinsic or extrinsic task of interest.

\section{Methodology}
\label{sec:method}
Before introducing the semi-supervised MTL approach to learning word meta-embeddings, we first outline the self-supervised learning baselines used in the comparisons.
First, we include both the aforementioned 1\texttt{TO}N/1\texttt{TO}N$^{+}$~\cite{yin2015learning} and standard AEs~\cite{bao2018meta} proposed in the prior work. 
CAEME concatenates the embedding set into a single vector and trains the AE to produce a lower-dimensional representation, while DAEME keeps the embedding vectors separate in the encoding. 

\subsection{Autoncoded Meta-Embedding}
We consider a dataset of $n$ samples, $D:=\{(\mat{x}_i, \vec{y}_i)\}^{n}$, where the input is represented as a word embedding input matrix $\mat{x} \in \mathbb{R}^{|V| \times d}$ with a corresponding target vector $\vec{y} \in \mathbb{R}^{|V|}$. Here $x_i$ is the embedding for a word $w 
\in V$ where $V$ denotes the vocabulary for $D$. Further $V \subset \mathcal{V}$ is a subset vocabulary for a given $D \in \mathcal{D}$ where $\mathcal{D}$ are all datasets and unique set of words for $\mathcal{D}$ is $\mathcal{V}$.

The matrix for all $D \in \mathcal{D}$ and $N$ embeddings in the ensemble set of word embeddings $X: {x_1, x_2, \ldots, x_N}$ can be expressed as a matrix $\mat{X} \in \mathbb{R}^{n \times k}$ which can be split into training matrix $\mat{X}_{tr} \in \mathbb{R}^{n_{tr} \times k}$ and test matrix $\mat{X}_{ts} \in \mathbb{R}^{n_{ts} \times k}$. We define $k=Nd$, $n_{tr}=|\mathcal{V}|-|V|$ to be the number of training examples (one $D$ is left out for training), $n_{ts}=|V|$ number of test examples, both obtained by row-wise concatenation of pretrained embeddings, as shown in Equation \ref{eq:concat_emb} and for simplicity, refer to $\mat{X}_{tr}$ as $\mat{X}$ herein.

\begin{equation}\label{eq:concat_emb}
\mat{X}:=\bigoplus_{i=1}^{N}\vec{x}_i
\end{equation}

In standard meta-embedding training, we obtain a hidden representation $\mat{Z} \in \mathbb{R}^{n_{tr} \times k}$ from $\mat{X}$ where $k \ll Nd$ using an AE $f_{\theta}$. 

\begin{equation}\label{eq:ae}
     \mat{Z}(w) = f_{\theta}(\vec{X}(w))
\end{equation}

In Section \ref{sec:method_ssmtl} we will discuss how $\mat{Z}$ is learned as an auxiliary task to the main supervised task which also shares $\mat{Z}$. We now consider reconstruction loss functions.

\paragraph{Meta-Embedding Loss Functions}
We compare training AEMEs with various loss functions against other meta-embedding approaches mentioned in related work such as 1TON~\cite{yin2015learning} and autoencoded meta-embeddings~\cite{bao2018meta}. This includes the MAE loss $\sum_{i=1}^{N}|\mat{X}_i - \vec{\hat{\mat{X}}_i}|$), MSE loss $\sum_{i=1}^{N}\big(\mat{X}_{i} - \hat{\mat{X}}_{i}\big)^{2}$ and the KL divergence. 
For minimizing the KL divergence, the AE output distributions for each sample in $\hat{\mat{X}}$ are normalized to form $\hat{p}(\mat{X})$ and the corresponding meta-embedding target distribution $p(\mat{X})$ using the \texttt{softmax} function.
The KL is then expressed as Equation \ref{eq:meta_kl} which corresponds to the reconstruction loss $\mathcal{L}_r$ shown in Equation \ref{eq:kl_loss}. Here, $M$ refers to the mini-batch size used for training. 

\begin{equation}\label{eq:meta_kl}
D_{\text{KL}}\Big(p(\mat{X})||p(\hat{\mat{X}})
\Big) = \sum_{i-1}^{M}p\big(\mat{X}_i\big)\log \frac{p(\hat{\mat{X}}_i)}{p(\mat{X}_i)}   
\end{equation}

\begin{equation}\label{eq:kl_loss}
   \mathcal{L}_{r}(\hat{\mat{X}},\mat{X})= \frac{1}{M}\sum_{i=1}^{M} 
   p(\mat{X}_{i}) \Bigg( \log \Big(p(\mat{X}_{i})\Big) - \log (\hat{p} \Big( \hat{\mat{X}}_{i} ) \Big) \Bigg)  
\end{equation}

Since $\tanh$ activations are used and input vectors are $\ell_2$ normalized we propose a Squared Cosine Proximity (SCP) loss, shown in Equation \ref{eq:cosine_loss}. This forces the optimization to tune weights such that the rotational difference between the embedding spaces is minimized, thus preserving semantic information in the reconstruction. In the context of its utility for the TAE, we also want to minimize the angular difference between corresponding vectors in different vector spaces. Unlike KL-divergence, the SCP loss is a proper distance metric since it is symmetric and satisfies the triangular inequality. 

\begin{equation}\label{eq:cosine_loss}
	\mathcal{L}_{r}(\hat{\mat{X}},\mat{X}) = \sum_{i=1}^{M}\Big(1 - \frac{\sum_{j=1}^{m}\hat{\mat{X}}_{ij} \cdot \mat{X}_{ij}}{\sqrt{\sum_{j=1}^{m} \hat{\mat{X}}_{ij}^{2}}\sqrt{\sum_{j=1}^{m}\mat{X}_{ij}^{2}}}\Big)^{2}
\end{equation}

\subsubsection{Model Configurations}

The AEME is a 1-hidden layer AE with a hidden layer dimension $d_h=200$. This is consistent for all tested loss functions, making for a fair performance comparison between the proposed SCP loss and KL divergence loss against MSE and MAE. We initialize all weights with a normal distribution of mean $\mu=0$ and standard deviation $\sigma=1$. The dropout rate is set to $p=0.2$ for all datasets. The model takes $|\mathcal{V}|= 4819$ unique vocabulary terms pertaining to all tested word association and word similarity datasets and performs Stochastic Gradient Descent (SGD) with a mini-batch size of 32 trained for 50 epochs for each dataset with an early stopping criteria. The hidden dimension size, batch size and number of epochs were chosen based on a small grid search. 

\subsection{Target Autoencoder Meta-Embedding}
We propose the Target AE (TAE), an AE variant that learns to predict a target embedding $\mat{X}^{t}_{tr} \in \mathbb{R}^{n_{tr} \times d}$,in the meta-embedding set, from the remaining source embeddings $\mat{X}^{s} \in \mathbb{R}^{n_{tr} \times d(N-1)}$. The AE $\mathtt{f}_{\Theta_i}: \mat{X}^{(s, i)}_{tr}$ $\to \mat{X}^{t}_{tr} \quad \forall i \in N$ permutations within the embedding set in a leave-one-out setting, where each $i$-th model is trained for each permutation. We then denote resulting meta-embedding as $\bar{Z}_i$ and $\mat{Z}:= \bar{\mat{Z}}$ where $k=d$ in our experiments for the TAE embedding. This embedding represents different combinations of mappings from one vector space to another. This is motivated by ~\cite{caruana1998multitask} who points out that treating inputs as auxiliary output tasks can be beneficial. This has also found in the success of large-scale language modelling where predicting a percentage (e.g 15\%) of masked tokens from unmasked tokens has led to better representations as measured on supervised tasks~\cite{devlin2018bert}.

In contrast, the TAE is similar to that of CAEME only the label is a single embedding from the embedding set and the input are remaining embeddings from the set. After training a TAE, the hidden layer encoding is concatenated with the original target vector. The Mean Target AutoEncoder (MTE) instead takes an average between different projections.

\subsection{Meta-Embedding for Supervised Task Regularization}\label{sec:method_ssmtl}

\subsubsection{MTL for Intrinsic Tasks}
\paragraph{Word Similarity}
The first instance of using meta-embeddings as an auxiliary reconstruction task is for learning word associations. Since, we are conbining self-supervised learning (reconstruction of the ensemble set) and supervised learning (word association scores) simultaneously, we view this approach as semi-supervised learning. The shared representation that is used during reconstruction is also shared as input for the main task.

\begin{figure}[ht]
\begin{center}
	\includegraphics[scale=0.45]{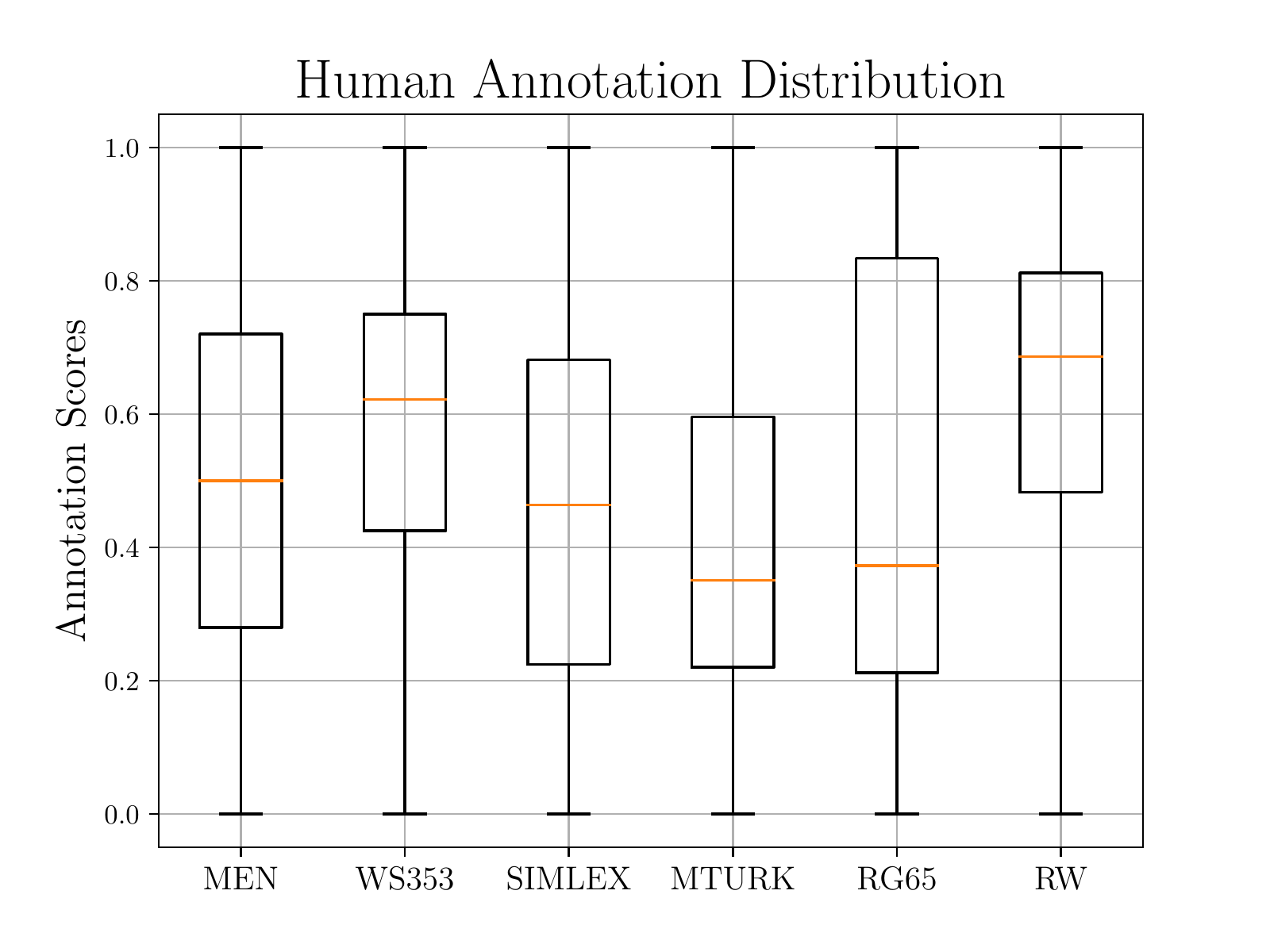}
 \caption{Word Association Annotation Distribution}\label{fig:annot_dist}
\end{center}
\end{figure}

The word similarity scores $\vec{y} \in [0,1]$ are normalized as different datasets are within different ranges. The resulting normalized $\vec{y}$ are considered as soft probabilistic targets. We also considered converting $\vec{y}$ to binary classes using a threshold from the mean $\bar{y}$. However, as illustrated in Figure \ref{fig:annot_dist}, the quartiles of the distribution over the annotation scores are not symmetric around the median, with the exception of MEN and Simlex. Moreover, factors such as annotation guidelines, part of speech (PoS) distribution and the concreteness of word pairs are different for each dataset. Such factors partially explain why the output distributions $\forall y \in \mathcal{Y}$ ($\mathcal{Y}$ corresponds to all outputs in $\mathcal{D}$) vary as seen in Figure \ref{fig:annot_dist}.

\begin{figure}[!bp]
\begin{center}
 \includegraphics[scale=0.6]{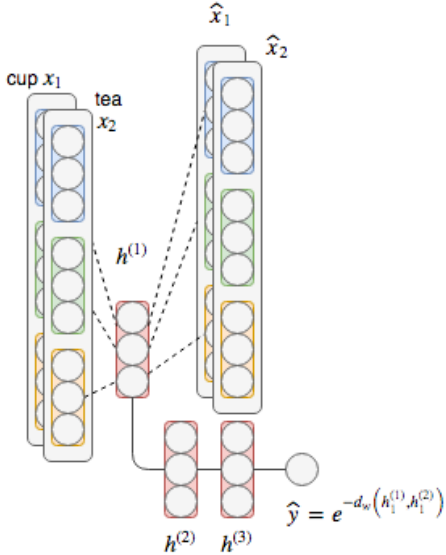}
 \caption{Multi-Task Meta-Embedding (an example of \textit{true} similarity where `teacup' and `teabag' are distinctly different.)}\label{fig:multi_task_meta}
\end{center}
\end{figure}

We train on all but one word similarity test dataset $D_{ts}$ using the AE to produce meta-embedding pairs $h^{1}$ for the word pair vectors that are also used on the test dataset as it is the unsupervised (self-supervised) learning part of the network. This is illustrated in Figure \ref{fig:multi_task_meta}, where red coloring indicates the hidden layer representations.

For training, the hidden layer dimension sizes are $200-50-10$, corresponding to $h^{(1)}$-$h^{(2)}$-$h^{(3)}$ in Figure \ref{fig:multi_task_meta}. For the layers that are only used on the main task ($h^{(2)}$ and ($h^{(3)}$) we denote their parameters as $\Theta$. This notation is also used to refer intrinsic task-specific layers. Furthermore we define the parameters of the main tasks final layer as $\omega \subset \Theta$.

Note that $h^{(1)}$ has dimensionality $d=200$, the same size as the aforementioned self-supervised learning approach that does not use MTL (ie word similarity is not learned). Once MTL has converged over a set of epochs and hyperparameters are tuned, we compare the $\rho_s$ of the shared hidden layer $h^{(1)}$ outputs, as opposed to using $\hat{y}$ produced in the siamese network that predicts word similarity directly. We test various distance measures for word similarity, including Manhattan, Euclidean and Cosine dissimilarity. 

Since the data is [0-1] normalized the pairwise distance can be computed as Equation \ref{eq:dist_metric} and is kept in this range using the negative exponent . This corresponds to estimating the probability density of the output targets, where $y$ are soft probabilistic targets. 

\begin{gather}\label{eq:dist_metric}
\hat{\vec{y}} = \exp\Big(-d_{\omega}(\mat{h}^{l}_{1}, \mat{h}^{l}_{2})\Big)
\end{gather}


The total loss $\mathcal{L} = \mathcal{L}_{r} + \mathcal{L}_{s}$ is then the sum of the reconstruction loss $\mathcal{L}_{r}$ shown in in Equation \ref{eq:recon_loss} and $\mathcal{L}_{s}$, the cross-entropy (CE) loss between predictions made by the meta-embedding shared representation shown in Equation \ref{eq:nll_mtl_loss}. In Equation $\ref{eq:recon_loss}$, $\lambda$ controls the amount of meta-embedding regularization during optimization.

\begin{align}\label{eq:recon_loss}
	\mathcal{L}_{r}(\mat{X}, \hat{\mat{X}}) = \frac{\lambda}{2M}\sum_{i=1}^{M}\sum_{j=1}^{2}\Big(\mat{X}_{ij}-\hat{\mat{X}}_{ij}\Big)^{2} \\
	\mathcal{L}_{s}(\vec{y}, \vec{\hat{y}})=\text{-}\frac{1}{M}\sum_{i=1}^{M} \vec{y}_i\log\vec{\hat{y}}_{i} + \big( 1-\vec{y}_{i} \big)\log \big(1-\vec{\hat{y}}_{i} \big)\label{eq:nll_mtl_loss}
\end{align}

We argue that when annotators decide on word similarity given a word pair that they choose on how $x_1$ relates to $x_2$ only, and not vice-versa~\cite{tversky1977features}. In other words, the relationship between two words is not strictly symmetric and the viewing order matters when humans are tasked with estimating word similarity. Therefore, when coming up with a similarity measure using this argument, we test an asymmetric similarity measure between $(h^{l}_2,h_{2}^{l})$ encodings. This simply involves replacing the denominator of the cosine similarity ($||x_{1}||_{2}||x_{2}||_{2}$) that is used as the distance function $d_{\omega}(\cdot)$ to $||x_{1}||_{2}\cdot ||x_{1}||_{2}$ before being passed to the negative exponent for the final probability output. Hence, the model is trained to learn how $x_1$ is related to $x_2$.

\paragraph{MTL for Analogy}
We also test our meta-embeddings as an auxiliary task for analogy, a task that plays a fundamental role in human cognition~\cite{gentner2001analogical,gentner2011computational}. Since not all algorithms used to train the pretrained embeddings are known to preserve analogies as a side effect, we might expect single embeddings for which this does hold to outperform meta-embeddings. We test if the word meta-embedding encodes analogical structure. Furthermore, we test if meta-embedding reconstruction improves performance. 

For analogy answers, we focus on CosAdd~\cite{mikolov2013linguistic} for measuring similarity between analogy pairs and ranking candidates accordingly. Hence, the meta-embedding scheme can be expressed as Equation \ref{eq:analog_form}, where $(w_a, w_b)$ are the first analogy pair and $(w_c, w_d)$ are the second, while $Z(w)$ is the meta-embedding representation for a given $w$.

{\small 
\begin{align}\label{eq:analog_form}
    \text{CosAdd}(\mat{Z}(w_a), \mat{Z}(w_b), \mat{Z}(w_c), \mat{Z}(w_d)) = \nonumber \\
    \text{Cos}(\mat{Z}(w_b) - \mat{Z}(w_a) + \mat{Z}(w_c), \mat{Z}(w_d))     
\end{align}
}
%


\subsubsection{MTL for Extrinsic Tasks}\label{sec:meta_downstream}
Finally, we test how such meta-embeddings perform in standard supervised learning problems in natural language tasks, more specifically, sequence prediction and text classification problems. This includes Named Entity Recognition (NER), Universal Dependency Part-of-Speech (UDPoS) tagging 

We learn to reconstruct the embeddings using the top performing autoencoding approach and use it to perform meta-embedding as an auxiliary task for log-likelihood training of one of the above sequence tagging problems. In this setting, $Y$ is a sequence of length $T$ with tokens $y_1, y_2,...y_T$, $\Theta$ are the parameters of an RNN with encoded hidden state vector $h_t = f(x_{t}, h_{t-1}; \Theta)$ and $P(y_t|h_t; \Theta)$ is the conditional computed using a linear projection followed by normalization, in our case, using a $\mathtt{softmax}$ function.

\begin{align}
\Theta^{*}= \argmax_{\Theta}\sum_{t=1}^{T}\log P(y_t|x_t, h_{t-1} ; \Theta)
\end{align}

The model uses tokens $x_{t}$ as the input and the hidden state $h_{t-1}$ to predict a tag $\hat{y}_{t-1}$. We find the optimal sequence of tags such that the likelihood of the predicted sequence of tags is maximised.

In the subsequent analysis, we learn to reconstruct the embeddings as in Equation \ref{eq:recon_loss} as an auxiliary task. However, instead of using two dense layers for predicting on the main task, as is the case for word similarity, we use the output of the shared representation as input to recurrent and convolutional layers in the aforementioned downstream tasks. We add a penalty $\lambda \mathcal{L}_r$, where the penalty coefficient $\lambda \in \mathbb{R}$ is tuned based on validation performance, effectively constraining optimization to also reconstruct the embedding set.

\section{Experiments}\label{sec:experiments}
The following source word embeddings are considered in the embedding set as they are publicly available and widely used for natural language tasks: $\mathtt{skipgram}$ and $\mathtt{cbow}$~\cite{mikolov2013efficient}, FastText~\cite{bojanowski2016enriching}, LexVec~\cite{salle2016matrix}, Hellinger PCA (HPCA)~\cite{lebret2013word} and Hierarchical Document Context (HDC)~\cite{sun2015learning}. 

\subsection{Intrinsic Evaluation}
\subsubsection{Word Similarity Results}

The following word association and word similarity datasets are used throughout experimentation: Simlex~\cite{hill2015simlex}, WordSim-353~\cite{finkelstein2001placing}, RG~\cite{rubenstein1965contextual}, MTurk (MechanicalTurk-771)~\cite{halawi2012large}, rare word (RW)~\cite{luong2014addressing} and MEN~\cite{bruni2012distributional}. 
Table \ref{tab:meta_results} shows the results, where (1) shows the single embedding performance, (2) results for standard meta-embedding approaches that either apply a single mathematical operation or use a linear projection as an encoding, (3) results using AE schemes by Bollegala and Bao~\cite{bao2018meta} and (4) results of our proposed TAE embedding. 
Results in red shading indicate the best performing meta-embedding for all presented approaches, while black shading indicates the best performing meta-embedding for the respective section. 

Best performing word meta-embeddings are held between CAEMEs that use the proposed Cosine-Embedding loss, while KL-divergence also performs well on Simlex and RW. Interestingly, both  Simlex and RW are distinct in that Simlex is the only dataset providing scores on \textit{true similarity} instead of free association, which has shown to be more difficult for word embeddings to account for~\cite{hill2016simlex}, while RW provides morphologically complex words to find similarity between. This suggests KL-divergence is well suited for encoding word relations that are relatively rare and perhaps more abstract relations (e.g Simlex contains less concrete terms in the vocabulary compared to other datasets, such as WS353). Similarly, we find SCP loss to achieve best results on RG and MEN, both the smallest and largest datasets of the set.

Furthermore, the TAE variant has lead to competitive performance against other meta-embedding approaches, showing good results on WS353. However, overall, standard AE performs better than the TAE. 

The AE that uses a squared cosine loss and a KL-divergence loss improves performance in the majority of the cases, reinforcing the argument that considering the angles explicitly through normalization (log-softmax for KL) is an important step in encoding large documents of varying length and semantics. Lastly, we have shown its use in the context of training word meta-embedding but cosine loss can also be used to minimize angular differences in standard word embedding training. 

\begin{table}
\centering
\captionsetup{justification=centering, margin=0cm}

\resizebox{0.95\linewidth}{!}{%
\begin{tabular}{c||ccccccc}
 \toprule[1.pt]
\multicolumn{1}{l}{1. Embeddings} & Simlex & WS353 & RG & MTurk & RW & MEN \\
\midrule

Skipgram & \cellcolor{black!20}44.19 & \cellcolor{red!20}\emph{77.17} & 76.08 & 68.15 & \cellcolor{black!20}49.70 & 75.85\\
FastText & 38.03 & 75.33 & 79.98 & 67.93 & 47.90 & 76.36 \\
GloVe & 37.05 & 66.24 & 76.95 & 63.32 & 36.69 & 73.75 \\
LexVec & 41.93 & 64.79 & 76.45 &  \cellcolor{black!20}71.15 & 48.94 & \cellcolor{black!20}80.92\\
HPCA & 16.60 & 57.11 & 41.72 & 37.45 & 13.36 & 34.90 \\
HDC & 40.68 & 76.81 & \cellcolor{black!20}80.58 & 65.76 & 46.34 & 76.03\\

\midrule
\multicolumn{1}{l}{2. Standard Meta} & & & & & & & \\
\midrule

CONC & \cellcolor{black!20}42.57 & \cellcolor{black!20}72.13 & \cellcolor{black!20}81.36 & \cellcolor{black!20}71.88 & 49.91 & 80.33 \\
SVD & 41.10 & 72.06 & 81.18 & 71.50 & 49.13 & 79.85 \\
AV & 40.63 & 70.50 & 80.05 & 70.51 & 49.28 & 78.31 \\
1TON & 41.30 & 70.19 & 80.20 & 71.52 & 50.80 & \cellcolor{black!20}80.39 \\
1TON* & 41.49 & 70.60 & 78.40 & 71.44 & \cellcolor{black!20}50.86 & 80.18\\

\midrule
\multicolumn{1}{l}{3. $\ell_2$-AE} & & & & & & & \\
\midrule

Decoupled & 42.56 & 70.62 & 82.81 & 71.16 & 50.79 & 80.33 \\
Concatenated & \cellcolor{black!20}43.10 & \cellcolor{black!20}71.69 & 84.52 & \cellcolor{red!20}\emph{71.88} & \cellcolor{black!20}50.78 & 81.18 \\

\midrule
\multicolumn{1}{l}{$\ell_1$-AE} & & & & & & & \\
\midrule

Decoupled & 43.52 & 70.30 & \cellcolor{black!20}82.91 & \cellcolor{black!20}71.43 & 51.48 & \cellcolor{black!20}81.16 \\
Concatenated & \cellcolor{black!20}44.41 & \cellcolor{black!20}70.96 & 81.16 & 69.63 & \cellcolor{black!20}51.89 & 80.92 \\

\midrule
\multicolumn{1}{l}{Cosine-AE} & & & & & & & \\
\midrule

Decoupled & 43.13 & 71.96 & 84.23 & 70.88 & 50.20 & 81.02 \\
Concatenated & \cellcolor{black!20}44.85 & \cellcolor{black!20}72.44 & \cellcolor{red!20}\emph{85.41} & 70.63 & \cellcolor{black!20}50.74 & \cellcolor{red!20}\emph{81.94} \\

\midrule
\multicolumn{1}{l}{KL-AE} & & & & & & & \\
\midrule
Decoupled & 44.13 & 71.96 & 84.23 & \cellcolor{black!20}70.88 & 50.20 & 81.02 \\
Concatenated & \cellcolor{red!20}\emph{45.10} & \cellcolor{black!20}74.02 & \cellcolor{black!20}85.34 & 67.75 & \cellcolor{red!20}\emph{53.02} & \cellcolor{black!20}81.14 \\

\midrule
\multicolumn{1}{l}{4. TAE +$Y$} & & & & & & & \\
\midrule


$\to$ Skipram & 42.43 & 75.33 & 80.11 & 66.51 & 44.77 & 78.98 \\
$\to$ FastText & 41.69 & 72.65 & 80.51 & 67.64 & 47.41 & 77.48  \\
$\to$ Glove & 41.75 & \cellcolor{black!20}76.65 & \cellcolor{black!20}82.40 & 68.92 & \cellcolor{black!20}48.83 & 78.27 \\
$\to$ LexVec & \cellcolor{black!20}42.85 & 73.33 & 80.97 & \cellcolor{black!20}69.17 & 46.71 & \cellcolor{black!20}79.63 \\
$\to$ HPCA & 40.03 & 69.65 & 70.43 & 61.31 & 36.38 & 73.10 \\
$\to$ HDC & 42.43 & 74.08 & 80.11 & 66.51 & 44.76 & 77.93  \\

\bottomrule[1.pt]
\end{tabular}%
}
 \caption{Meta-Embedding Unsupervised Results ($\rho_s$)}
  \label{tab:meta_results}
\end{table}

Table \ref{tab:meta_results} shows the results for the self-supervised learning methods, where grey represents the best model for each section (1-4) and red represents the best model for all sections (same for proceeding tables). It is clearly difficult to obtain relatively good performance on Simlex and RW. The former was introduced to make a clear distinction between association and true similarity, hence the annotation scores reflect this difference, making it difficult for DSMs which solely rely on word associations. In contrast, we see in Table \ref{tab:multi_meta_results} there is a large improvement in $\rho_{s}$ over these datasets using SS-MTL. In experimentation, we found performance with a 2-hidden layer network was similar to a single layer network. Given the sample size of $|\mathcal{V}|=4819$ for the word similarity, we are not surprised that a relatively smaller sample size relies less on a deeper representation.

The first measure (e.g Cosine-) represents the reconstruction loss $\mathcal{L}_{r}$ and second represents the word similarity loss $\mathcal{L}_{s}$ (e.g Binary CE). A cosine $\mathcal{L}_{r}$ and the Brier's score $\mathcal{L}_{s}$\footnote{Brier's score~\cite{brier1950verification} is a score between probabilistic predictions and is equivalent to MSE for regression.} are found to perform the best on average.
Since word similarity scores are not directly optimized when using maximum likelihood, it is not obvious this is a suitable objective for improving on the evaluation metric $\rho_s$. Therefore, we also consider Brier's score, which can be considered as MSE for class probabilities. Indeed, in Table \ref{tab:multi_meta_results} we find that using Brier's score for the annotations improves $\rho_s$ for 4 of 6 datasets. 

Meta-embeddings that are learned only using unsupervised methods (Equation \ref{tab:meta_results}) give $\rho_{s} = 45.10$, on Simlex, while the semi-supervised MTL approach produces the most noticeable performance gain with a dramatic increase of $\rho_{s} = 74.73$. Although datasets such as Simlex have made a clear distinction between word associations and true similarity, we find there is still performance improvements made on true similarity when transferring knowledge in the form of meta-embeddings from different annotation distributions that only score word association and not the true similarity ~\cite{hill2016simlex}.

\begin{table}
\centering
\captionsetup{justification=centering, margin=0cm}

\resizebox{1.0\linewidth}{!}{%
\begin{tabular}{c||ccccccc}
 \toprule[1.pt]
\multicolumn{1}{l}{} & Simlex & WS353 & RG & MTurk & RW & MEN \\

\midrule
Cosine-OLS & 53.63 & 73.13 & 83.07 & 69.41 & 60.49 & 80.25 \\
Cosine-NLL & 59.22 & 76.09 & 80.45 & 70.43 & 61.31 & 82.49  \\
Cosine-Brier's & \cellcolor{black!20}63.72 & \cellcolor{red!20}\emph{80.21} & \cellcolor{red!20}\emph{89.54} & \cellcolor{red!20}\emph{83.45} & \cellcolor{black!20}\emph{70.76} & \cellcolor{red!20}\emph{84.14}  \\

\midrule
$\ell_1$-OLS  & 55.16 & 68.80 & 82.82 & 70.35 & 61.07 & 78.56\\
$\ell_1$-NLL & 53.54 & 77.82 & 82.09 & 73.12 & 64.46 & 79.12  \\
$\ell_1$-Brier's & \cellcolor{black!20}68.78 & \cellcolor{black!20}77.60 & \cellcolor{black!20}87.44 & \cellcolor{black!20}80.67 & \cellcolor{black!20}78.05 & \cellcolor{black!20}79.73  \\

\midrule
$\ell_2$-OLS & 68.31 & \cellcolor{black!20}73.85 & 84.48 & 70.91 & 53.20 & \cellcolor{black!20}81.60  \\
$\ell_2$-NLL & 53.80 & 71.15 & 85.10 & 71.51 & 50.61 & 79.38 \\
$\ell_2$-Brier's & \cellcolor{red!20}\emph{74.73} & 69.68 & \cellcolor{black!20}85.29 & \cellcolor{black!20}76.30 & \cellcolor{red!20}\emph{80.07} & 70.64   \\

\midrule
KL-OLS & 62.47 & \cellcolor{black!20}68.93 & 85.75 & 72.35 & 50.38 & 80.95  \\
KL-NLL  & 48.91 & 67.93 & 86.67 & 72.33 & 48.91 & 78.98 \\
KL-Brier's & \cellcolor{black!20}71.39 & 66.91 & \cellcolor{black!20}87.58 & \cellcolor{black!20}73.43 & \cellcolor{black!20}67.11 & \cellcolor{black!20}81.78  \\

\bottomrule[1.pt]
\end{tabular}%
}

\caption{Semi-Supervised Multi-Task Word Embedding Learning ($\rho_s$) Results on Word Similarity}
  \label{tab:multi_meta_results}
\end{table}
In the semi-supervised MTL setting shown in Table \ref{tab:multi_meta_results}, we see that results are also consistent with Table \ref{tab:meta_results} as the cosine loss in reconstruction results in best performance for 4 out of the 6 datasets.

\subsubsection{Analogy Results}
We evaluate how the learned models from Table \ref{tab:multi_meta_results} transfer to analogy tasks, namely MSR Word Representation dataset~\cite{gao2014wordrep} (8000 questions with 8 relations), Google Analogy dataset~\cite{mikolov2013distributed} (19,544 questions with 15 relations) amd SemEval 2012 Task 2 Competition Dataset~\cite{jurgens2012semeval} (3218 question with 79 relations). The former two consist of categories of different analogy questions and the latter includes ranked candidate word pairs based on word pair relational similarity for a set of chosen word pairs.  
CosAdd~\cite{mikolov2013linguistic} is used for calculating the analogy answers for Google and MSR which ranks candidates given as CosAdd$(a, b, c, d) = \cos(b-a+c, d)$ and chooses the answer as the highest ranking candidate. 


Table \ref{tab:analogy_results} shows the results of transferring the learned semi-supervised multi-task learning (SS-MTL) embeddings to analogy tasks. Here, we analyse (1) whether the word meta-embeddings carry over to analogy even if not all embedding algorithms preserve analogy relations, (2) check if the similarity encoded with SS-MTL has any effect on performance on analogy and (3) check whether the nonlinearity induced by autoencoding, performs relatively well (somewhat counter-intuitively, given the existence of linear relationships between analogy pairs~\cite{allen2019analogies}) for analogy reasoning. 

In general, SS-MTL that incorporates similarity scores has some transferability to analogy based on the scores provided by the aforementioned word similarity datasets.
For Google Analogy, the larger of the three datasets with the smallest range of relation types, we find that the SS-MTL model that previously trained with Cosine-Brier's loss functions shows the best performance overall. This is consistent with findings from Table 2 where the same model performs best for 4 of 6 word similarity datasets. This suggests that performing additional nonlinear meta-word encoding somewhat preserves the linear structures preserved in models such as skipgram and fasttext. Additionally, we find Brier's score to perform particularly well based on $\rho_s$ results.

\begin{table}
\centering
\captionsetup{justification=centering, margin=0cm}

\resizebox{.7\linewidth}{!}{%
\begin{tabular}{c||ccc}
 \toprule[1.pt]
\multicolumn{1}{l}{} & MSR & Google & SemEval \\

\midrule
Skipgram & 73.13 & 72.89 & 22.64\\
FastText & \cellcolor{black!20}64.19 & 73.82 & 24.77\\
GloVe &  71.45 & 71.73 & \cellcolor{black!20}19.98 \\
LexVec & 74.03 & 67.28 & 21.49\\

\midrule
Cosine-OLS & 73.24 & 71.57 & 22.13 \\
Cosine-NLL & 71.23 & 68.39 &  20.16\\
Cosine-Brier's & 74.78\cellcolor{black!20} & \emph{74.18}\cellcolor{red!20} & 23.44 \\

\midrule
$\ell_1$-OLS  & 69.32 & 68.21 & 20.45 \\
$\ell_1$-NLL & 68.69 & 67.27 & 19.02  \\
$\ell_1$-Brier's & \cellcolor{black!20}70.37 & \cellcolor{black!20}72.55 & \cellcolor{black!20} 20.36 \\

\midrule
$\ell_2$-OLS & 73.20 & 72.16 & 22.71 \\
$\ell_2$-NLL & 72.37 & 69.35 & 21.08  \\
$\ell_2$-Brier's & \emph{75.72}\cellcolor{red!20} & 74.11 & \cellcolor{red!20}\emph{24.84} \\

\midrule
KL-OLS & 68.08\cellcolor{black!20} & 65.28 & 18.24\\
KL-NLL & 65.51 & 65.90 & 19.66 \\
KL-Brier's & 64.30 & 67.22 & \cellcolor{black!20}20.75 \\

\bottomrule[1.pt]

\end{tabular}%
}
\caption{SS-MTL Embedding Analogy Transferability}
\label{tab:analogy_results}
\end{table}

\begin{table*}
	\centering
	\def\arraystretch{1.0}
	\begin{small}
	\resizebox{0.9\linewidth}{!}{%
	\begin{tabular}{lc|cc|cc|cc|cc|cc}
		\toprule[1.pt]

		\textbf{Task}  & \textbf{Model} &   \multicolumn{2}{c}{\textbf{Skipgram}} & \multicolumn{2}{c}{\textbf{FastText}} & 	\multicolumn{2}{c}{\textbf{GloVe}} & \multicolumn{2}{c}{\textbf{LexVec}} & \multicolumn{2}{c}{\textbf{Meta}} \\
        \midrule
        \parbox[t]{2mm}{\multirow{5}{*}{\rotatebox[origin=c]{90}{\textbf{NER}}}}\\

		& CNN & 81.25 & 79.77 & 83.19 & 81.46 & 84.62 & 81.93 & 80.08 & 79.24 & 88.15 & 86.63 \\
		& LSTM & 83.59 & 81.61 & 84.80 & 82.27 & 82.29 & 80.08 & 82.65 & 80.74 & \cellcolor{black!20}91.23 & \cellcolor{black!20}\emph{88.58} \\
		& GRU & 82.14 & 80.59 & 83.94 & 82.08 & 84.72 & 81.80 & 83.19 & 81.47 & 90.79 & 88.38 \\
		& Highway & 81.72 & 80.39 & 83.88 & 82.61 & 82.17 & 81.28 & 84.55 & 81.24 & 89.85 & 87.05 \\
		
		\midrule

	    \parbox[t]{2mm}{\multirow{5}{*}{\rotatebox[origin=c]{90}{\textbf{UDPOS}}}}\\
		
		& CNN & 88.78 & 87.42 & 88.91 & 87.57 & 88.76 & 87.49 & 87.18 & 87.01 & 89.43 & 88.38 \\
		& LSTM   & 90.43 & 89.55 & 90.61 & 88.91 & 90.64 & 89.62 & 90.39 & 89.16 & \cellcolor{black!20}91.89 & \cellcolor{black!20}\emph{91.12} \\
		& GRU   & 89.86 & 88.23 & 89.24 & 88.44 & 89.17 & 88.23 & 89.02 & 88.17 & 90.72 & 90.59 \\
		& Highway & 90.11 & 88.73 & 89.97 & 88.46 & 89.81 & 88.02 & 89.23 & 88.08 & 90.26 & 90.01 \\
		
		\midrule

		\parbox[t]{2mm}{\multirow{5}{*}{\rotatebox[origin=c]{90}{\textbf{Sentiment}}}}\\
		
		& CNN & \multicolumn{2}{c|}{84.03} & \multicolumn{2}{c|}{85.42} &  \multicolumn{2}{c|}{83.73} & \multicolumn{2}{c|}{84.44} & \multicolumn{2}{c}{89.21}\\
		& LSTM & \multicolumn{2}{c|}{86.50} & \multicolumn{2}{c|}{87.21} & \multicolumn{2}{c|}{86.48} & \multicolumn{2}{c|}{85.17} &\multicolumn{2}{c}{\cellcolor{black!20}\emph{\textbf{92.38}}} \\
		& GRU & \multicolumn{2}{c|}{85.43} & \multicolumn{2}{c|}{87.69} & \multicolumn{2}{c|}{84.73} & \multicolumn{2}{c|}{85.01}  & \multicolumn{2}{c}{91.75} \\
		& Highway   & \multicolumn{2}{c|}{82.39} & \multicolumn{2}{c|}{85.31} & \multicolumn{2}{c|}{84.21} & \multicolumn{2}{c|}{84.58} & \multicolumn{2}{c}{89.73} \\

		\bottomrule[1pt]
	\end{tabular}%
	}
	\end{small}
	\caption{Validation (left) \& Test (right) Accuracy (\%): NER, UDPoS \& Sentiment Analysis (only test acc.)  \iffalse and Chunking tasks.\fi}

	\label{tab:downstream_results}
\end{table*}

\subsection{Extrinsic Evaluation}
For \textbf{NER} we use the New York Times NER recipe tagging task~\cite{greene2015extracting} that contains 17.5k recipes which accumulates to 67.5k steps, 142.5k tags.
We use 16,622 sentences  for \textbf{Universal Dependency PoS}~\cite{nivre2016universal} (254k words) that contain weblogs, newsgroups, emails and reviews, which are used to define universal dependencies for the English UD Treebank (15/02/2017 version 2.0). The trees are converted to Stanford Dependencies and manually corrected to universal dependencies, predominantly by single annotations.


IMDB Movie reviews dataset~\cite{maas2011learning} is used for \textbf{Sentiment Analysis} where we predict positive and negative sentiments for 50k reviews, where both positive and negative classses are balanced and the train/test split is also 50-50. All reviews are filtered to have at least 30 reviews, reviews were assigned as positive if the average score is 7/10 or higher, negative if $\leq 4$ and neutral reviews are discarded. 

\subsubsection{Downstream Task Results} 

Table \ref{tab:downstream_results} shows the results on all 3 tasks, comparing the performance of each single pretrained embedding in the ensemble set to word meta-embeddings (Meta) that uses the reconstruction as an auxiliary task, as mentioned in  \autoref{sec:method_ssmtl}.
Based on the validation performance we set $\lambda = 0.1$ for NER and UDPoS and $\lambda= 0.15$ for Sentiment Analysis. Unlike the intrinsic tasks, we found a 2-hidden layer AE improved sentence-level meta embedding reconstruction for Sentiment Analysis. This is the only sentence classification dataset and therefore we use sentence-level meta-embedding regularization as opposed to word-level meta-embeddings that are used for intrinsic tasks and the remaining extrinsic tasks (NER and UDPoS tagging).

\begin{figure}[ht]
\begin{center}
 \includegraphics[scale=0.45]{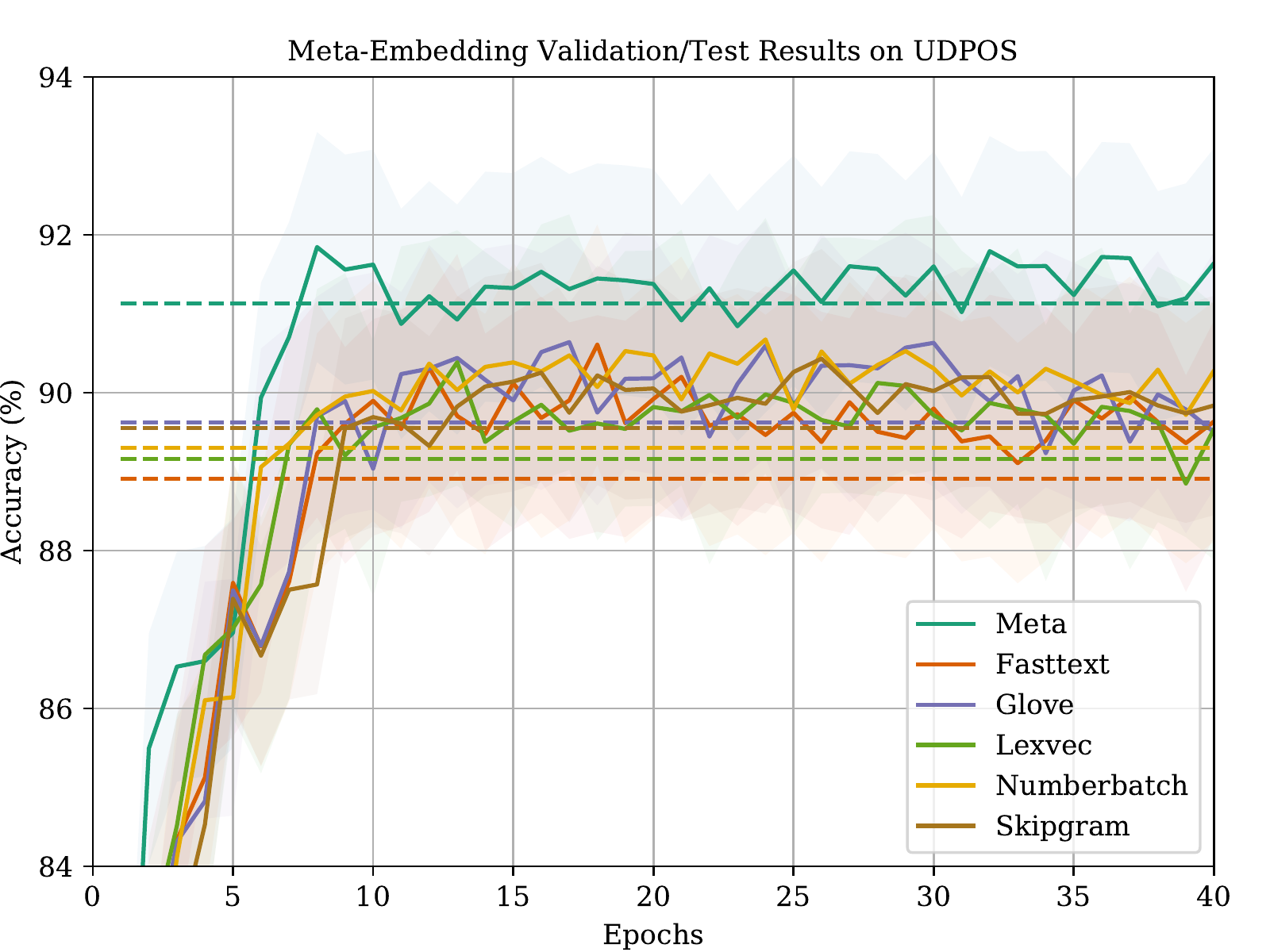}
 \caption{LSTM Results with Multi-Task Meta-Embedding on Universal Dependency PoS Tagging}\label{fig:udpos_meta_results}
\end{center}
\end{figure}

Improvements are found using meta-embedding reconstruction as an auxiliary task, when compared to using single pretrained embeddings. Overall, best results are obtained using meta-embeddings with an LSTM for the main task, while remaining models all show an increase in validation and test accuracy.
Figure \ref{fig:udpos_meta_results} shows results for UDPoS tagging where meta-embeddings have led to better results than using pretrained embeddings  in multi-task learning. We find that near convergence the accuracy deviation in the validation set is lowered when using meta-embeddings, improving stability and calibrated probability estimates on the supervised tasks.  This is also found for NER and sentiment classification across all models tested. Moreover, meta-embedding reconstruction greatly improves the performance early on ($<$10 epochs), which means that less training time is needed when the shared layer is forced to preserve information from the embedding set.

\section{Conclusion}
This paper proposed a multi-task learning approach to learning word meta-embeddings as an auxiliary reconstruction task to improve predictions on a main intrinsic (e.g word similarity, analogy) or extrinsic (e.g PoS, NER) task whereby the meta-embedding layer is a shared representation between tasks. 
We find consistent improvements against baselines and also identify objective functions for meta-embedding reconstruction that lead to optimal performance on the main task. In doing so, we identified a meta-embedding target autoencoder that learns to project between different permutations of different embedding spaces within the ensemble set and use the the mean of the resulting latent representations as the meta-embedding.

We find performance increased significantly when using manually annotated scores from word similarity datasets in comparison to single word embeddings and unsupervised word meta-embedding approaches. We also find that angular-based loss functions are well suited for word meta-learning for both self-supervised learning and the proposed multi-task semi-supervised learning method, showing best results on 4 out of the 6 word similarity datasets in both cases.
Most significant improvements were found on relatively difficult word similarity and association datasets such as Simlex and rare word, while still improving by a large margin on the remaining datasets. Moreover, we find slight improvements made when transferring the semi-supervised models for analogy tasks.  
Lastly, we find consistent improvement when using meta-embeddings as an auxiliary task for downstream tasks such as Named Entity Recognition, Sentiment Classification and Universal Dependency PoS tagging.

However, this is expected given that similarity scores are more general than specific word pair relation types and not all word embedding algorithms preserve analogical relations to the same degree.

\bibliography{ecai}
\end{document}